\documentclass[11pt,a4paper]{article}
\usepackage[hyperref]{emnlp2018}
\usepackage{times}
\usepackage{latexsym}
\usepackage{graphicx}

\usepackage{color}
\usepackage{booktabs}
\usepackage{graphicx}
\usepackage{amsmath}

\newcommand{\todo}[1]{\textcolor{blue}{[TODO: #1]} }

\usepackage{url}

\aclfinalcopy 

\title{Causal Explanation Analysis on Social Media}

\author{
  Youngseo Son, Nipun Bayas, and H. Andrew Schwartz \\
  Stony Brook University \\
  Stony Brook, NY \\
  {\tt \{yson,nbayas,has\}@cs.stonybrook.edu} \\}

\date{}

\begin{document}
\maketitle
\begin{abstract}
Understanding causal explanations --- reasons given for happenings in one's life --- has been found to be an important psychological factor linked to physical and mental health. 
Causal explanations are often studied through manual identification of phrases over limited samples of personal writing.
Automatic identification of causal explanations in social media, while challenging in relying on contextual and sequential cues, offers a larger-scale alternative to expensive manual ratings and opens the door for new applications (e.g. studying prevailing beliefs about causes, such as climate change). 
Here, we explore automating causal explanation analysis, building on  discourse parsing, and presenting two novel subtasks:
\textit{causality detection} (determining whether a causal explanation exists at all) and \textit{causal explanation identification} (identifying the specific phrase that is the explanation).
We achieve strong accuracies for both tasks but find different approaches best: an SVM for causality prediction ($F1 = 0.791$) and a hierarchy of Bidirectional LSTMs for \textit{causal explanation identification} ($F1 = 0.853$). 
Finally, we explore applications of our complete pipeline ($F1 = 0.868$), showing demographic differences in mentions of causal explanation and that the association between a word and sentiment can change when it is used within a causal explanation. 
\end{abstract}

\section{Introduction}

Explanations of happenings in one's life, \textit{causal explanations}, are an important topic of study in social, psychological, economic, and behavioral sciences. 
For example, psychologists have analyzed people's causal explanatory style~\cite{peterson1988pessimistic} and  found strong negative relationships with  depression, passivity, and hostility, as well as positive relationships with life satisfaction, quality of life, and length of life~\cite{scheier1989dispositional,carver1987optimism,peterson1988pessimistic}.

\begin{figure}
\centering
  \includegraphics[width=3.0in]{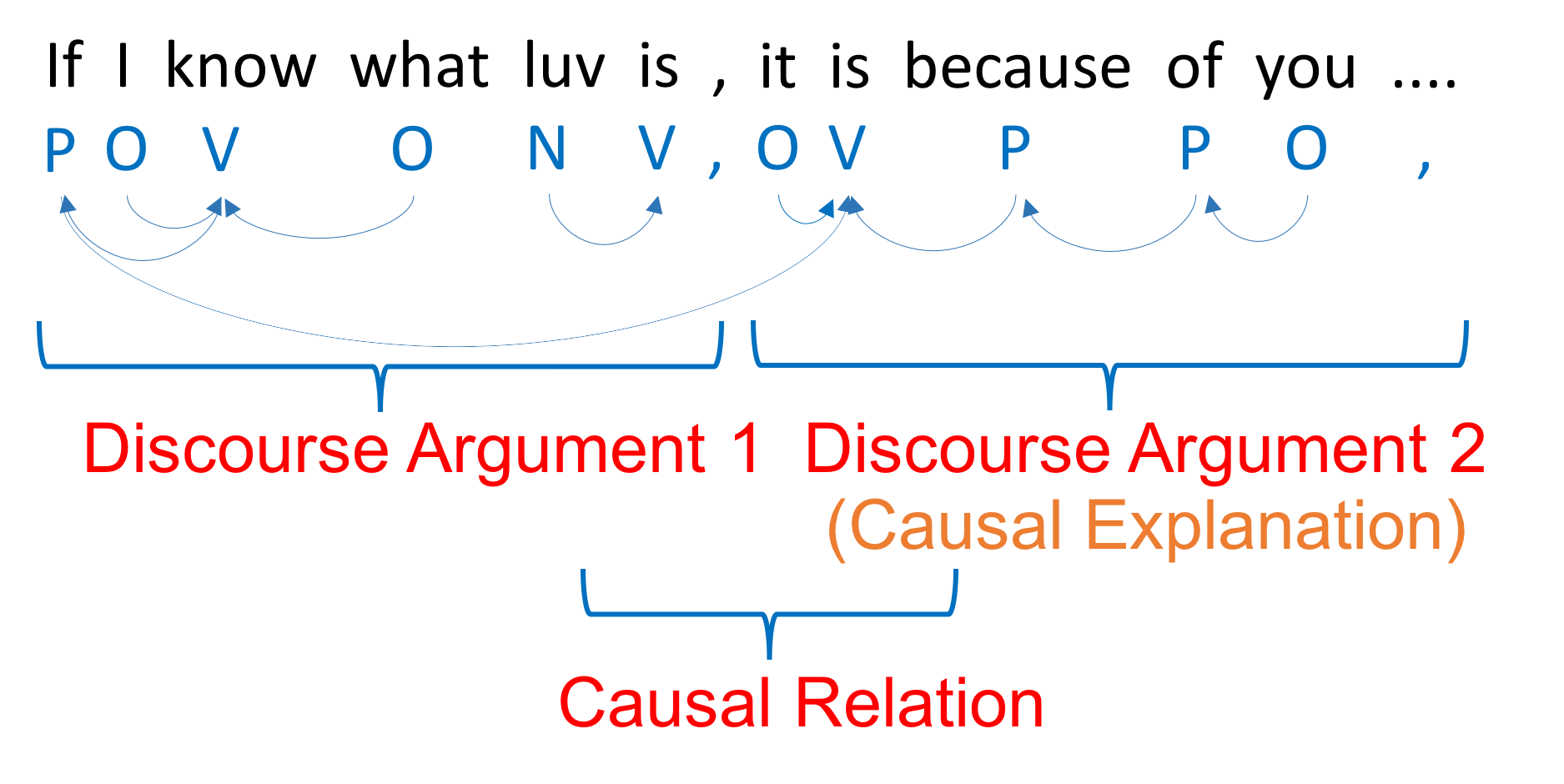}
  \caption{A casual relation characterizes the connection between two discourse arguments, one of which is the \textit{causal explanation}. }
  \label{fig:causal_relation_parsing}
\end{figure}

To help understand the significance of causal explanations, consider how they are applied to measuring optimism (and its converse, pessimism)~\cite{peterson1988pessimistic}. 
For example, in ``My parser failed \textit{because I always have bugs.}'', the emphasized text span is considered a causal explanation which indicates pessimistic personality -- a negative event where the author believes the cause is pervasive.
However, in ``My parser failed \textit{because I barely worked on the code.}'', the explanation would be considered a signal of optimistic personality -- a negative event for which the cause is believed to be short-lived.

Language-based models which can detect causal explanations from everyday social media language can be used for more than automating optimism detection. 
Language-based assessments would enable other large-scale downstream tasks: tracking prevailing causal beliefs (e.g., about climate change or autism), better extracting process knowledge from non-fiction (e.g., gravity causes objects to move toward one another), or detecting attribution of blame or praise in product or service reviews (``I loved this restaurant because the fish was cooked to perfection''). 

In this paper, we introduce causal explanation analysis and its subtasks of detecting the presence of causality (\textit{causality prediction}) and identifying explanatory phrases (\textit{causal explanation identification}). There are many challenges to achieving these task. First, the ungrammatical texts in social media incur poor syntactic parsing results which drastically affect the performance of  discourse relation parsing pipelines \footnote{Off-the-shelf Penn Discourse Treebank (PDTB) end-to-end parsers perform poorly on our Facebook causal prediction dataset (see Table~\ref{tab:causality})}. Many causal relations are \textit{implicit} and do not contain any discourse markers (e.g., `because').  Further, \textit{Explicit} causal relations are also more difficult in social media due to the abundance of abbreviations and variations of discourse connectives (e.g., `cuz' and `bcuz').

Prevailing approaches for social media analyses, utilizing traditional linear models or bag of words models (e.g., SVM trained with n-gram, part-of-speech (POS) tags, or lexicon-based features) alone do not seem appropriate for this task since they simply cannot segment the text into meaningful discourse units or discourse arguments \footnote{Each discourse relation theory uses a different term for minimal discourse text spans: `Elementary Discourse Unit (EDU)' in RST and `Discourse Argument' in PDTB. We will call it `Discourse Argument' in this paper, since we adapted the PDTB text segmentation method.} such as clauses or sentences rather than random consecutive token sequences or specific word tokens. 
Even when the discourse units are clear, parsers may still fail to accurately identify discourse relations since the content of social media is quite different than that of newswire which is typically used for discourse parsing. 

In order to overcome these difficulties of discourse relation parsing in social media, we simplify and minimize the use of syntactic parsing results and capture relations between discourse arguments, and investigate the use of a recursive neural network model (RNN). 
Recent work has shown that RNNs are effective for utilizing discourse structures for their downstream tasks \cite{ji2017neural,bhatia2015better,wieting2015towards, paulus2014global}, but they have yet to be directly used for discourse relation prediction in social media.
We evaluated our model by comparing it to off-the-shelf end-to-end discourse relation parsers and traditional models. We found that the SVM and random forest classifiers work better than the LSTM classifier for the causality detection, while the LSTM classifier outperforms other models for identifying causal explanation.

The contributions of this work include: (1) the proposal of models for both (a) causality prediction and (b) causal explanation identification, (2) the extensive evaluation of a variety of models from social media classification models and discourse relation parsers to RNN-based application models, demonstrating that feature-based models work best for causality prediction while RNNs are superior for the more difficult task of causal explanation identification,  (3) performance analysis on architectural differences of the pipeline and the classifier structures, (4) exploration of the applications of causal explanation to downstream tasks, and (5) release of a novel, anonymized causality Facebook dataset along with our causality prediction and causal explanation identification models.

\section{Related Work}
Identifying causal explanations in documents can be viewed as discourse relation parsing. 
The Penn Discourse Treebank (PDTB) \cite{prasad2007penn} has a `Cause' and `Pragmatic Cause' discourse type under a general `Contingency' class and Rhetorical Structure Theory (RST) \cite{mann1987rhetorical} has a `Relations of Cause'. In most cases, the development of discourse parsers has taken place \textit{in-domain}, where researchers have used the existing annotations of discourse arguments in newswire text (e.g. Wall Street Journal) from the discourse treebank and focused on exploring different features and optimizing various types of models for predicting relations \cite{pitler2009automatic,park2012improving,zhou2010predicting}. 
In order to further develop automated systems, researchers have proposed end-to-end discourse relation parsers, building models which are trained and evaluated on the annotated PDTB and RST Discourse Treebank (RST DT). 
These corpora consist of documents from Wall Street Journal (WSJ) which are much more well-organized and grammatical than social media texts \cite{biran:15,lin2014pdtb,ji2014representation,feng2014linear}. 

Only a few works have attempted to parse discourse relations for out-of-domain problems such as text categorizations on social media texts; Ji and Bhatia used models which are pretrained with RST DT for building discourse structures from movie reviews, and Son adapted the PDTB discourse relation parsing approach for capturing counterfactual conditionals from tweets \cite{bhatia2015better,ji2017neural,son2017recognizing}. 
These works had substantial differences to what propose in this paper. 
First, Ji and Bhatia used a pretrained model (not fully optimal for some parts of the given task) in their pipeline; Ji's model performed worse than the baseline on the categorization of legislative bills, which is thought to be due to legislative discourse structures differing from those of the training set (WSJ corpus). Bhatia also used a pretrained model finding that utilizing discourse relation features did not boost accuracy~\cite{bhatia2015better,ji2017neural}. 
Both Bhatia and Son used manual schemes which may limit the coverage of certain types of positive samples-- Bhatia used a hand-crafted schema for weighting discourse structures for the neural network model and Son manually developed seven surface forms of counterfactual thinking for the rule-based system~\cite{bhatia2015better,son2017recognizing}. 
We use social-media-specific features from pretrained models which are directly trained on tweets and we avoid any hand-crafted rules except for those included in the existing discourse argument extraction techniques. 

The automated systems for discourse relation parsing involve multiple subtasks from segmenting the whole text into discourse arguments to classifying discourse relations between the arguments. 
Past research has found that different types of models and features yield varying performance for each subtask. 
Some have optimized models for discourse relation classification (i.e.~given a document indicating if the relation existing) without discourse argument parsing using models such as Naive-Bayes or SVMs, achieve relatively stronger accuracies but a simpler task than that associated with discourse arguments~\cite{park2012improving,zhou2010predicting,pitler2009automatic}.
Researchers who, instead, tried to build the end-to-end parsing pipelines considered a wider range of approaches including sequence models and RNNs~\cite{biran:15,feng2014linear,ji2014representation,li2014recursive}. Particularly, when they tried to utilize the discourse structures for out-domain applications, they used RNN-based models and found that those models are advantageous for their downstream tasks \cite{bhatia2015better,ji2017neural}.

In our case, for identifying causal explanations from social media using discourse structure, we build an RNN-based model for its structural effectiveness in this task (see details in section \ref{sec:rnn_model}). However, we also note that simpler models such as SVMs and logistic regression obtained the state-of-the-art performances for text categorization tasks in social media \cite{lynn2017human,mohammad2013nrc}, so we build relatively simple models with different properties for each stage of the full pipeline of our parser.
\section{Methods}
We build our model based on PDTB-style discourse relation parsing since PDTB has a relatively simpler text segmentation method;\footnote{RST parsing builds fully hierarchical discourse tree structures out of the whole span of target text which highly depends on syntactic parsing and exact matching of elementary discourse units which are extremely hard to obtain from social media texts} for explicit discourse relations, it finds the presence of discourse connectives within a document and extracts discourse arguments which parametrize the connective while for implicit relations, it considers all adjacent sentences as candidate discourse arguments.
\subsection{Dataset}

We created our own causal explanation dataset by collecting 3,268 random Facebook status update messages. Three well-trained annotators manually labeled whether or not each message contains the causal explanation and obtained 1,598 causality messages with substantial agreement ($\kappa=0.61$). We used the majority vote for our gold standard. Then, on each causality message, annotators identified which text spans are causal explanations. 

For each task, we used 80\% of the dataset for training our model and 10\% for tuning the hyperparameters of our models. Finally, we evaluated all of our models on the remaining 10\% (Table~\ref{tab:causality_dataset} and Table~\ref{tab:causality_tweet_da}). For causal explanation detection task, we extracted discourse arguments using our parser and selected discourse arguments which most cover the annotated causal explanation text span as our gold standard.  

\begin{table}
\centering
\begin{tabular}{|l|l|l|l|}
\hline \bf Dataset & Causality & Non-Causal & Total   \\ \hline
Training & 1,284 & 1,330  & 2,614  \\
Validation & 150  & 177 & 327  \\
Test & 164 & 163 & 327 \\ \hline
Total & 1,598 & 1,670 & 3,268 \\
\hline
\end{tabular}
\caption{\label{tab:causality_dataset} Number of messages containing causality or not in our dataset. }
\end{table}

\begin{table}
\centering
\begin{tabular}{|l|l|l|l|}
\hline \bf Causality messages & CE DA & Total DA   \\ \hline
Training & 1,278   & 5,606  \\
Validation & 160 & 652  \\
Test & 160 & 757 \\ \hline
Total & 1,598 & 7,015 \\
\hline
\end{tabular}
\caption{\label{tab:causality_tweet_da} The number of discourse arguments in causality messages. Across 1,598 total causality messages, we found 7,015 discourse arguments (Total DA) and the one which covers annotated causal explanation are used as causal explanation discourse arguments (CE DA)}
\end{table}

\subsection{Model}
We build two types of models. First, we develop feature-based models which utilize features of the successful models in social media analysis and causal relation discourse parsing. Then, we build a recursive neural network model which uses distributed representation of discourse arguments as this approach can even capture latent properties of causal relations which may exist between distant discourse arguments. We specifically selected bidirectional LSTM since the model with the discourse distributional structure built in this form outperformed the traditional models in similar NLP downstream tasks \cite{ji2017neural}.

\paragraph{Discourse Argument Extraction}
As the first step of our pipeline, we use Tweebo parser \cite{kong2014dependency} to extract syntactic features from messages. Then, we demarcate sentences using punctuation (`,') tag and periods. Among those sentences, we find discourse connectives defined in PDTB annotation along with a Tweet POS tag for conjunction words which can also be a discourse marker. In order to decide whether these connectives are really discourse connectives (e.g., I went home, \textbf{but} he stayed) as opposed to simple connections of two words (I like apple \textbf{and} banana) we see if verb phrases \footnote{minimal discourse unit is verb phrases with very few exceptions \cite{prasad2007penn}} exist before and after the connective by using dependency parsing results. Although discourse connective disambiguation is a complicated task which can be much improved by syntactic features \cite{pitler2009using}, we try to minimize effects of syntactic parsing and simplify it since it is highly error-prone in social media. Finally, according to visual inspection, emojis (`E' tag) are crucial for discourse relation in social media so we take them as separate discourse arguments (e.g.,in ``My test result... :('' the sad feeling is caused by the test result, but it cannot be captured by plain word tokens).

\begin{figure*}
  \centering
  \includegraphics[width=4.8in]{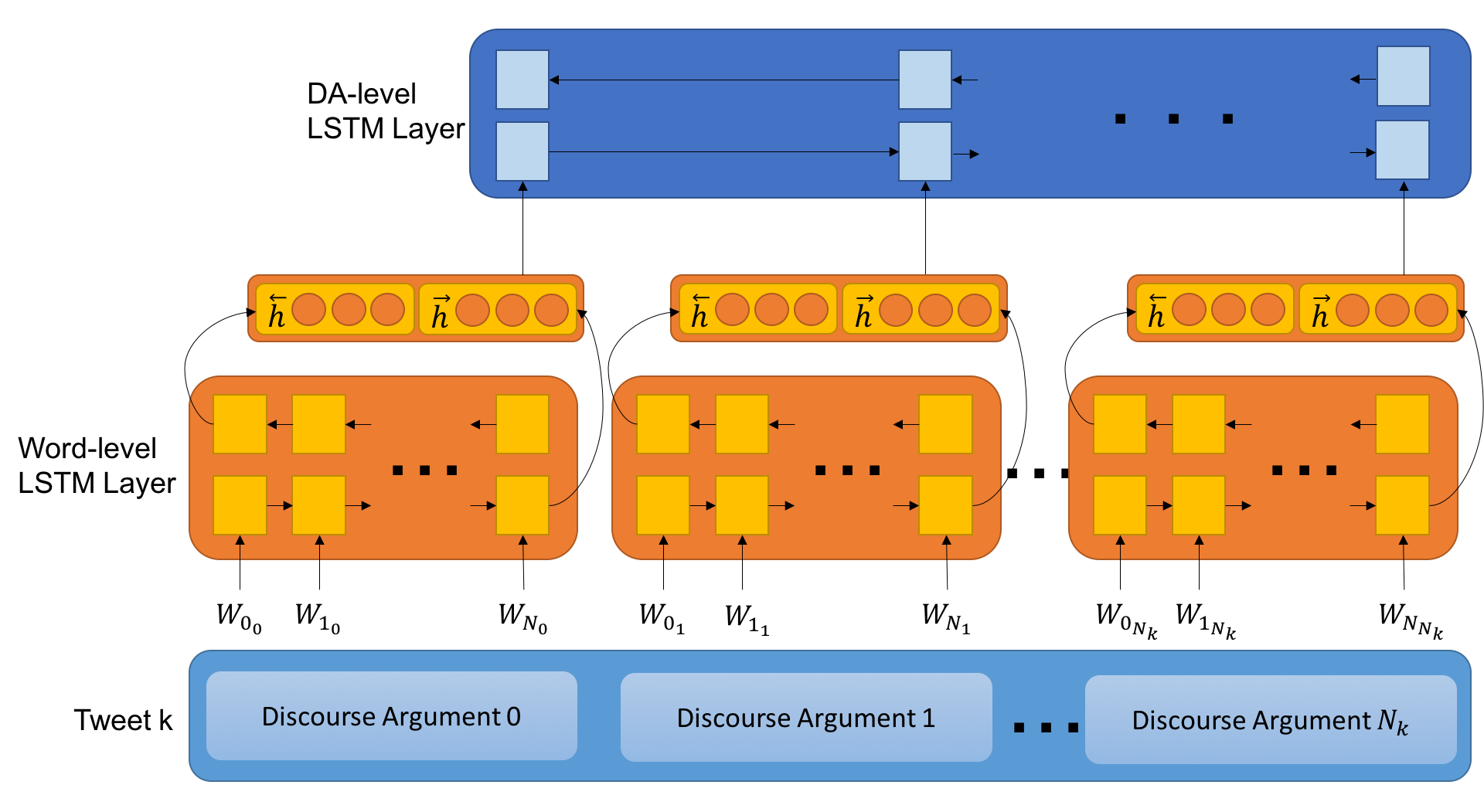}
  \caption{LSTM classifier for causality detection and explanation identification}
  \label{fig:LSTM_classifier}
\end{figure*}

\paragraph{Feature Based Models}

We trained a linear SVM, an rbf SVM, and a random forest with N-gram, charater N-gram, and tweet POS tags, sentiment tags, average word lengths and word counts from each message as they have a pivotal role in the models for many NLP downstream tasks in social media \cite{mohammad2013nrc,lynn2017human}. In addition to these features, we also extracted \textit{First-Last, First3}  features and \textit{Word Pairs} from every adjacent pair of discourse arguments since these features were most helpful for causal relation prediction \cite{pitler2009automatic}. \textit{First-Last, First3} features are first and last word and first three words of two discourse arguments of the relation, and \textit{Word Pairs} are the cross product of words of those discourse arguments. These two features enable our model to capture interaction between two discourse arguments. \cite{pitler2009automatic} reported that these two features along with verbs, modality, context, and polarity (which can be captured by N-grams, sentiment tags and POS tags in our previous features) obtained the best performance for predicting Contingency class to which causality belongs.

\paragraph{\label{sec:rnn_model} Recursive Neural Network Model}
We load the GLOVE word embedding \cite{pennington2014glove} trained in Twitter \footnote{\url{http://nlp.stanford.edu/data/glove.twitter.27B.zip}} for each token of extracted discourse arguments from messages. For the distributional representation of discourse arguments, we run a Word-level LSTM on the words' embeddings within each discourse argument and concatenate last hidden state vectors of forward LSTM ($\overrightarrow{h}$) and backward LSTM ($\overleftarrow{h}$) which is suggested by \cite{ji2017neural} ($DA = [\overrightarrow{h};\overleftarrow{h}]$). Then, we feed the sequence of the vector representation of discourse arguments to the Discourse-argument-level LSTM (DA-level LSTM) to make a final prediction with log softmax function. With this structure, the model can learn the representation of interaction of tokens inside each discourse argument, then capture discourse relations across all of the discourse arguments in each message (Figure \ref{fig:LSTM_classifier}). In order to prevent the overfitting, we added a dropout layer between the Word-level LSTM and the DA-level LSTM layer.

\paragraph{Architectural Variants}
We also explore subsets of the full RNN architecture, specifically with one of the two LSTM layers removed. 
In the first model variant, we directly input all word embeddings of a whole message to a BiLSTM layer and make prediction (\textbf{Word LSTM}) without the help of the distributional vector representations of discourse arguments. In the second model variant, we take the average of all word embeddings of each discourse argument ($DA_k=\frac{1}{N_k} \sum_{i=1}^{N_k}W_{i}$), and use them as inputs to a BiLSTM layer (\textbf{DA AVG LSTM})  as the average vector of embeddings were quite effective for representing the whole sequence  \cite{ji2017neural,wieting2015towards}.
As with the full architectures, for CP both of these variants ends with a many-to-one classification per message, while the CEI model ends with a sequence of classifications. 

\subsection{Experiment}
\paragraph{Feature Based Model}
We explored three types of models (RBF SVM, Linear SVM, and Random Forest Classifier) which have previously been shown empirically useful for the language analysis in social media. We filtered out low frequency \textbf{Word Pairs} features as they tend to be noisy and sparse \cite{pitler2009automatic}. Then, we conducted univariate feature selection to restrict all remaining features to those showing at least a small relationship with the outcome. Specifically, we keep all features passing a family-wise error rate of $\alpha = 60$ with the given outcome. After comparing the performance of the optimized version of each model, we also conducted a feature ablation test on the best model in order to see how much each feature contributes to the causality prediction. 

\paragraph{Neural Network Model}
We used bidirectional LSTMs for causality classification and causal explanation identification since the discourse arguments for causal explanation can show up either before and after the effected events or results and we want our model to be optimized for both cases. However, there is a risk of overfitting due to the dataset which is relatively small for the high complexity of the model, so we added a dropout layer (p=0.3) between the Word-level LSTM and the DA-level LSTM.

For tuning our model, we explore the dimensionality of word vector and LSTM hidden state vectors of discourse arguments of 25, 50, 100, and 200 as pretrained GLOVE vectors were trained in this setting. For optimization, we used Stochastic Gradient Descent (SGD) and Adam \cite{kingma2014adam} with learning rates 0.01 and 0.001.

We ignore missing word embeddings because our dataset is quite small for retraining new word embeddings. However, if embeddings are extracted as separate discourse arguments, we used the average of all vectors of all discourse arguments in that message. Average embeddings have performed well for representing text sequences in other tasks~\cite{wieting2015towards}.

\paragraph{Model Evaluation}
We first use state-of-the-art PDTB taggers for our baseline \cite{lin2014pdtb,biran:15} for the evaluation of the causality prediction of our models (\cite{biran:15} requires sentences extracted from the text as its input, so we used our parser to extract sentences from the message). Then, we compare how models work for each task and disassembled them to inspect how each part of the models can affect their final prediction performances. We conducted McNemar's test to determine whether the performance differences are statistically significant at $p < .05$. 
\section{Results}
We investigated various models for both causality detection and explanation identification. Based on their performances on the task, we analyzed the relationships between the types of models and the tasks, and scrutinized further for the best performing models. For performance analysis, we reported weighted F1 of classes.

\subsection{Causality Prediction}
\begin{table}
\centering
\begin{tabular}{|l|l|l|l|}
\hline \bf Model & F1  \\ \hline
\cite{biran:15} & 0.434 \\
\cite{lin2014pdtb} & 0.638\\ \hline
Linear SVM & \textbf{0.791}  \\
RBF SVM & \textbf{0.777} \\
Random Forest & 0.771\\ \hline
LSTM & 0.758\\

\hline
\end{tabular}
\caption{\label{tab:causality} Causality prediction performance across different predictive models. Bold indicates significant improvement over the LSTM}
\end{table}

\begin{table}
\centering
\begin{tabular}{|l|l|l|l|}
\hline \bf Model & F1 \\ \hline
All & 0.791 \\ 
\hline
- First-Last, First3 & 0.788 \\ 
- Word Pairs & 0.787 \\
- POS tags & 0.734\\
- (Char + Word) N-grams & 0.769 \\
- Sentiment tags & 0.791 \\
\hline
\end{tabular}
\caption{\label{tab:lsvm-feat}Feature ablation test of Linear SVM for causality prediction}
\end{table}

In order to classify whether a message contains causal relation, we compared off-the-shelf PDTB parsers, linear SVM, RBF SVM, Random forest and LSTM classifiers. The off-the-shelf parsers achieved the lowest accuracies (\cite{biran:15} and \cite{lin2014pdtb} in Table \ref{tab:causality}). This result can be expected since 1) these models were trained with news articles and 2) they are trained for all possible discourse relations in addition to causal relations (e.g., contrast, condition, etc). Among our suggested models, SVM and random forest classifier performed better than LSTM and, in the general trend, the more complex the models were, the worse they performed. This suggests that the models with more direct and simpler learning methods with features might classify the causality messages better than the ones more optimized for capturing distributional information or non-linear relationships of features.

\paragraph{Causality Classifier Analysis}
Table \ref{tab:lsvm-feat} shows the results of a feature ablation test to see how each feature contributes to causality classification performance of the linear SVM classifier. 
POS tags caused the largest drop in F1. 
We suspect POS tags played a unique role because discourse connectives can have various surface forms (e.g., because, cuz, bcuz, etc) but still the same POS tag `P'.
Also POS tags can capture the occurrences of modal verbs, a feature previously found to be very useful for detecting similar discourse relations~\cite{pitler2009automatic}. N-gram features caused 0.022 F1 drop while sentiment tags did not affect the model when removed. Unlike the previous work where \textit{First-Last, First3} and \textit{Word pairs} tended to gain a large F1 increase for multiclass discourse relation prediction, in our case, they did not affect the prediction performance compared to other feature types such as POS tags or N-grams. 

\subsection{Causal Explanation Identification}
\begin{table}
\centering
\begin{tabular}{|l|l|l|l|}
\hline \bf Model & Prec & Rec & F1  \\ \hline
Linear SVM & 0.773 & 0.727 & 0.743\\
RBF SVM & 0.739 & 0.771 & 0.749 \\
Random Forest & 0.747 & 0.790 & 0.746 \\
\hline
LSTM & \textbf{0.851} & \textbf{0.858} & \textbf{0.853} \\
\hline
\end{tabular}
\caption{\label{tab:ce} Causal explanation identification performance. Bold indicates significant imrpovement over next best model ($p < .05$)}
\end{table}

In this task, the model identifies causal explanations given the discourse arguments of the causality message. We explored over the same models as those we used for causality (sans the output layer), and found the almost opposite trend of performances (see Table \ref{tab:ce}). The Linear SVM obtained lowest F1 while the LSTM model made the best identification performance. As opposed to the simple binary classification of the causality messages, in order to detect causal explanation, it is more beneficial to consider the relation across discourse arguments of the whole message and implicit distributional representation due to the implicit causal relations between two distant arguments.

\subsection{Architectural Variants}

\begin{table}
\centering
\begin{tabular}{|l|l|l|l|}
\hline \bf Model & CP (F1) & CEI (F1) \\ 
\hline
Full LSTM & 0.758 & 0.853\\ 
\hline
DA AVG LSTM & 0.685 & 0.818 \\
Word LSTM & 0.694 & 0.792 \\
\hline
\end{tabular}

\caption{\label{tab:lstm-variant} The effect of Word-level LSTM (Word LSTM) and discourse argument LSTM (DA AVG LSTM) for causality prediction (CP) and causal explanation identification (CEI). Note that, as described in methods, there are architectural differences for CP and CEI models with the same names, most notably that the output layer is always a single classification for CP and a sequence of classifications for CEI. }
\end{table}

For causality prediction, we experimented with only word tokens in the whole message without help of Word-level LSTM layer (\textbf{Word LSTM}), and F1 dropped by 0.064  (CP in Table \ref{tab:lstm-variant}). Also, when we used the average of the sequence of word embeddings of each discourse argument as an input to the DA-level LSTM and it caused F1 drop of 0.073. This suggests that the information gained from both the interaction of words in and in between discourse arguments help when the model utilizes the distributional representation of the texts.

For causal explanation identification, in order to test how the LSTM classifier works without its capability of capturing the relations between discourse arguments, we removed DA-level LSTM layer and ran the LSTM directly on the word embedding sequence for each discourse argument for classifying whether the argument is causal explanation, and the model had 0.061 F1 drop (\textbf{Word LSTM} in CEI in Table \ref{tab:lstm-variant}). Also, when we ran DA-level LSTM on the average vectors of the word sequences of each discourse argument of messages, F1 decreased to 0.818. This follows the similar pattern observed from other types of models performances (i.e., SVMs and Random Forest classifiers) that the models with higher complexity for capturing the interaction of discourse arguments tend to identify causal explanation with the higher accuracies.

For CEI task, we found that when the model ran on the sequence representation of discourse argument (\textbf{DA AVG LSTM}), its performance was higher than the plain sequence of word embeddings (\textbf{Word LSTM}). Finally, in both subtasks, when the models ran on both Word-level and DA-Level (\textbf{Full LSTM}), they obtained the highest performance.  

\subsection{Complete Pipeline}

\begin{table}
\centering
\begin{tabular}{|l|l|l|l|}
\hline \bf Model & Prec & Rec & F1 \\
\hline
CP + CEI$_{causal}$ & \bf 0.864 & \bf 0.877 & \textbf{0.868} \\ 
CP + CEI$_{all}$ & 0.842 & 0.864 & 0.848 \\ 
CEI$_{causal}$ Only & 0.847 & 0.788 & 0.810 \\
CEI$_{all}$ Only & 0.836 & 0.848 & 0.842 \\ 
\hline
\end{tabular}
\caption{\label{tab:pipeline} The effect of Linear SVM Cauality model (CP) within our pipeline. CEI$_{all}$: LSTM CEI models trained on all messages; CEI$_{causal}$: LSTM CEI models trained only on causality messages (CEI$_{causal}$); CP + CEI$_{all|causal}$: the combination of Linear SVM and each LSTM model. Bold: significant ($p < .05$) increase in F1 over the next best model, suggesting the two-step approach worked best. }
\end{table}

Evaluations thus far zeroed-in on each subtask of causal explanation analysis (i.e.~CEI only focused on data already identified to contain causal explanations). Here, we seek to evaluate the complete pipeline of CP and CEI, starting from all of test data (those or without causality) and evaluating the final accuracy of CEI predictions. 
This is intended to evaluate CEI performance under an applied setting where one does not already know whether a document has a causal explanation. 

There are several approaches we could take to perform CEI starting from unannotated data. 
We could simply run CEI prediction by itself (\textbf{CEI Only}) or the pipeline of CP first and then only run CEI on documents predicted as causal (\textbf{CP + CEI}). 
Further, the CEI model could be trained only on those documents annotated causal (as was done in the previous experiments) or on all training documents including many that are not causal. 

Table \ref{tab:pipeline} show results varying the pipeline and how CEI was trained. Though all setups performed decent ($F1 > 0.81$) we see that the pipelined approach, first predicting causality (with the linear SVM) and then predicting causal explanations only for those with marked causal (CP + CEI$_{causal}$) yielded the strongest results. 
This also utilized the CEI model only trained on those annotated causal. 
Besides performance, an added benefit from this two step approach is that the CP step is less computational intensive of the CEI step and approximately 2/3 of documents will never need the CEI step applied.

\paragraph{Limitations.} We had an inevitable limitation on the size of our dataset, since there is no other causality dataset over social media and the annotation required an intensive iterative process. This might have limited performances of more complex models, but considering the processing time and the computation load, the combination of the linear model and the RNN-based model of our pipeline obtained both the high performance and efficiency for the practical applications to downstream tasks. In other words, it's possible the linear model will not perform as well if the training size is increased substantially. However, a linear model could still be used to do a first-pass, computationally efficient labeling, in order to shortlist social media posts for further labeling from an LSTM or more complex model. 

\section{Exploration}

Here, we explore the use of causal explanation analysis for downstream tasks. 
First we look at the relationship between use of causal explanation and one's demographics: age and gender. Then, we consider their use in sentiment analysis for extracting the causes of polarity ratings. Research involving human subjects was approved by the University of Pennsylvania Institutional Review Board. 


\paragraph*{Demographic differences.}

We first explored variance in number of causality posts by demographics. 
To do this, we used self-authored posts from a random 300 consenting-users of the MyPersonality dataset~\cite{kosinski2013private}.
For each user we calculate a $cp\_ratio$, defined as the number of causality predicted posts divided by their total number of posts, indicating the percentage of their posts which include a causal explanation. 
We then correlated this ratio with real-valued age using Pearson correlation and looked the differences by dichotomous gender using Cohen's \textit{d} (the difference in standardized means; only binary gender was available). We found significant ($p < .05$) moderate-sized associations for both, indicating both older individuals and females were likely to use more causal explanations. 

\paragraph*{Causality in Sentiment Analysis}
We explored the application of causality explanation identification for sentiment analysis using the Yelp polarity dataset~\cite{zhang2015character}. We randomly selected 10,000 of both positive and negative reviews and ran our complete pipeline on them to extract the causal explanations from the reviews. We then analyzed the ngrams from (a) causal explanation and (b) all other discourse arguments testing for associations with polarity. We used the a Bayesian interpretation of the log odds ratio using an informative dirichlet prior defined by~\newcite{monroe2008fightin}. We found difference in the top ngrams depending on whether the argument the ngram originated from was a causal explanation or not  (see Table \ref{tab:yelp}).  Top ngrams for causal explanations included more content words (e.g.~`rude', `overpriced', `slow') suggesting analyzing causal explanations within reviews can better target the \textit{reasons} for the negative review.

\begin{table}
\centering
\begin{tabular}{|l|c|c|}
\hline & \textbf{CE} & \textbf{Non-CE}\\
 & Top Ngrams & Top Ngrams \\
\hline
1 & worst & not \\ 
2 & was &  no \\ 
3 & not &  "   \\ 
4 & the worst & asked  \\ 
5 & horrible & she \\ 
6 & rude & told  \\ 
7 & bad  & said  \\ 
8 & overpriced  & minutes  \\ 
9 & over & ? \\ 
10 & slow &  me  \\ 
\hline
\end{tabular}
\caption{\label{tab:yelp} Top words most associated with negative reviews from within causal explanations (CE) and outside of causal explanation (Non-CE). }
\end{table}


\section{Conclusion}
We developed a pipeline for causal explanation analysis over social media text, including both causality prediction and causal explanation identification. We  examined a variety of model types and RNN architectures for each part of the pipeline, finding an SVM best for causality prediction and a hierarchy of BiLSTMs for causal explanation identification, suggesting the later task relies more heavily on sequential information. 
In fact, we found replacing either layer of the hierarchical LSTM architecture (the word-level or the DA-level) with a an equivalent ``bag of features'' approach resulted in reduced accuracy. 
Results of our whole pipeline of causal explanation analysis were found quite strong, achieving an $F1=0.868$ at identifying discourse arguments that are causal explanations. 

Finally, we demonstrated use of our models in applications, finding associations between demographics and rate of mentioning causal explanations, as well as showing differences in the top words predictive of negative ratings in Yelp reviews.
Utilization of discourse structure in social media analysis has been a largely untapped area of exploration, perhaps due to its perceived difficulty.
We hope the strong results of causal explanation identification here leads to the integration of more syntax and deeper semantics into social media analyses and ultimately enables new applications beyond the current state of the art.

\section*{Acknowledgments}
This work was supported, in part, by a grant from the Templeton Religion Trust (ID \#TRT0048). The funders had no role in study design, data collection and analysis, decision to publish, or preparation of the manuscript. We also thank Laura Smith, Yiyi Chen, Greta Jawel and Vanessa Hernandez for their work in identifying causal explanations.
\bibliography{emnlp2018}

\begin{thebibliography}{27}
\expandafter\ifx\csname natexlab\endcsname\relax\def\natexlab#1{#1}\fi

\bibitem[{Bhatia et~al.(2015)Bhatia, Ji, and Eisenstein}]{bhatia2015better}
Parminder Bhatia, Yangfeng Ji, and Jacob Eisenstein. 2015.
\newblock Better document-level sentiment analysis from rst discourse parsing.
\newblock \emph{arXiv preprint arXiv:1509.01599}.

\bibitem[{Biran and McKeown(2015)}]{biran:15}
Or~Biran and Kathleen McKeown. 2015.
\newblock Pdtb discourse parsing as a tagging task: The two taggers approach.
\newblock In \emph{Proceedings of the 16th Annual Meeting of the Special
  Interest Group on Discourse and Dialogue}, pages 96--104.

\bibitem[{Carver and Gaines(1987)}]{carver1987optimism}
Charles~S Carver and Joan~Gollin Gaines. 1987.
\newblock Optimism, pessimism, and postpartum depression.
\newblock \emph{Cognitive therapy and Research}, 11(4):449--462.

\bibitem[{Feng and Hirst(2014)}]{feng2014linear}
Vanessa~Wei Feng and Graeme Hirst. 2014.
\newblock A linear-time bottom-up discourse parser with constraints and
  post-editing.
\newblock In \emph{Proceedings of the 52nd Annual Meeting of the Association
  for Computational Linguistics (Volume 1: Long Papers)}, volume~1, pages
  511--521.

\bibitem[{Ji and Eisenstein(2014)}]{ji2014representation}
Yangfeng Ji and Jacob Eisenstein. 2014.
\newblock Representation learning for text-level discourse parsing.
\newblock In \emph{ACL (1)}, pages 13--24.

\bibitem[{Ji and Smith(2017)}]{ji2017neural}
Yangfeng Ji and Noah Smith. 2017.
\newblock Neural discourse structure for text categorization.
\newblock \emph{arXiv preprint arXiv:1702.01829}.

\bibitem[{Kingma and Ba(2014)}]{kingma2014adam}
Diederik~P Kingma and Jimmy Ba. 2014.
\newblock Adam: A method for stochastic optimization.
\newblock \emph{arXiv preprint arXiv:1412.6980}.

\bibitem[{Kong et~al.(2014)Kong, Schneider, Swayamdipta, Bhatia, Dyer, and
  Smith}]{kong2014dependency}
Lingpeng Kong, Nathan Schneider, Swabha Swayamdipta, Archna Bhatia, Chris Dyer,
  and Noah~A Smith. 2014.
\newblock A dependency parser for tweets.

\bibitem[{Kosinski et~al.(2013)Kosinski, Stillwell, and
  Graepel}]{kosinski2013private}
Michal Kosinski, David Stillwell, and Thore Graepel. 2013.
\newblock Private traits and attributes are predictable from digital records of
  human behavior.
\newblock \emph{Proceedings of the National Academy of Sciences},
  110(15):5802--5805.

\bibitem[{Li et~al.(2014)Li, Li, and Hovy}]{li2014recursive}
Jiwei Li, Rumeng Li, and Eduard Hovy. 2014.
\newblock Recursive deep models for discourse parsing.
\newblock In \emph{Proceedings of the 2014 Conference on Empirical Methods in
  Natural Language Processing (EMNLP)}, pages 2061--2069.

\bibitem[{Lin et~al.(2014)Lin, Ng, and Kan}]{lin2014pdtb}
Ziheng Lin, Hwee~Tou Ng, and Min-Yen Kan. 2014.
\newblock A pdtb-styled end-to-end discourse parser.
\newblock \emph{Natural Language Engineering}, 20(02):151--184.

\bibitem[{Lynn et~al.(2017)Lynn, Son, Kulkarni, Balasubramanian, and
  Schwartz}]{lynn2017human}
Veronica Lynn, Youngseo Son, Vivek Kulkarni, Niranjan Balasubramanian, and
  H~Andrew Schwartz. 2017.
\newblock Human centered nlp with user-factor adaptation.
\newblock In \emph{Proceedings of the 2017 Conference on Empirical Methods in
  Natural Language Processing}, pages 1146--1155.

\bibitem[{Mann and Thompson(1987)}]{mann1987rhetorical}
William~C Mann and Sandra~A Thompson. 1987.
\newblock \emph{Rhetorical structure theory: A theory of text organization}.
\newblock University of Southern California, Information Sciences Institute.

\bibitem[{Mohammad et~al.(2013)Mohammad, Kiritchenko, and
  Zhu}]{mohammad2013nrc}
Saif~M Mohammad, Svetlana Kiritchenko, and Xiaodan Zhu. 2013.
\newblock Nrc-canada: Building the state-of-the-art in sentiment analysis of
  tweets.
\newblock \emph{arXiv preprint arXiv:1308.6242}.

\bibitem[{Monroe et~al.(2008)Monroe, Colaresi, and Quinn}]{monroe2008fightin}
Burt~L Monroe, Michael~P Colaresi, and Kevin~M Quinn. 2008.
\newblock Fightin'words: Lexical feature selection and evaluation for
  identifying the content of political conflict.
\newblock \emph{Political Analysis}, 16(4):372--403.

\bibitem[{Park and Cardie(2012)}]{park2012improving}
Joonsuk Park and Claire Cardie. 2012.
\newblock Improving implicit discourse relation recognition through feature set
  optimization.
\newblock In \emph{Proceedings of the 13th Annual Meeting of the Special
  Interest Group on Discourse and Dialogue}, pages 108--112. Association for
  Computational Linguistics.

\bibitem[{Paulus et~al.(2014)Paulus, Socher, and Manning}]{paulus2014global}
Romain Paulus, Richard Socher, and Christopher~D Manning. 2014.
\newblock Global belief recursive neural networks.
\newblock In \emph{Advances in Neural Information Processing Systems}, pages
  2888--2896.

\bibitem[{Pennington et~al.(2014)Pennington, Socher, and
  Manning}]{pennington2014glove}
Jeffrey Pennington, Richard Socher, and Christopher~D. Manning. 2014.
\newblock Glove: Global vectors for word representation.
\newblock In \emph{Empirical Methods in Natural Language Processing (EMNLP)},
  pages 1532--1543.

\bibitem[{Peterson et~al.(1988)Peterson, Seligman, and
  Vaillant}]{peterson1988pessimistic}
Christopher Peterson, Martin~E Seligman, and George~E Vaillant. 1988.
\newblock Pessimistic explanatory style is a risk factor for physical illness:
  a thirty-five-year longitudinal study.
\newblock \emph{Journal of personality and social psychology}, 55(1):23.

\bibitem[{Pitler et~al.(2009)Pitler, Louis, and Nenkova}]{pitler2009automatic}
Emily Pitler, Annie Louis, and Ani Nenkova. 2009.
\newblock Automatic sense prediction for implicit discourse relations in text.
\newblock In \emph{Proceedings of the Joint Conference of the 47th Annual
  Meeting of the ACL and the 4th International Joint Conference on Natural
  Language Processing of the AFNLP: Volume 2-Volume 2}, pages 683--691.
  Association for Computational Linguistics.

\bibitem[{Pitler and Nenkova(2009)}]{pitler2009using}
Emily Pitler and Ani Nenkova. 2009.
\newblock Using syntax to disambiguate explicit discourse connectives in text.
\newblock In \emph{Proceedings of the ACL-IJCNLP 2009 Conference Short Papers},
  pages 13--16. Association for Computational Linguistics.

\bibitem[{Prasad et~al.(2007)Prasad, Miltsakaki, Dinesh, Lee, Joshi, Robaldo,
  and Webber}]{prasad2007penn}
Rashmi Prasad, Eleni Miltsakaki, Nikhil Dinesh, Alan Lee, Aravind Joshi, Livio
  Robaldo, and Bonnie~L Webber. 2007.
\newblock The penn discourse treebank 2.0 annotation manual.

\bibitem[{Scheier et~al.(1989)Scheier, Matthews, Owens, Magovern, Lefebvre,
  Abbott, and Carver}]{scheier1989dispositional}
Michael~F Scheier, Karen~A Matthews, Jane~F Owens, George~J Magovern, R~Craig
  Lefebvre, R~Anne Abbott, and Charles~S Carver. 1989.
\newblock Dispositional optimism and recovery from coronary artery bypass
  surgery: the beneficial effects on physical and psychological well-being.
\newblock \emph{Journal of personality and social psychology}, 57(6):1024.

\bibitem[{Son et~al.(2017)Son, Buffone, Raso, Larche, Janocko, Zembroski,
  Schwartz, and Ungar}]{son2017recognizing}
Youngseo Son, Anneke Buffone, Joe Raso, Allegra Larche, Anthony Janocko, Kevin
  Zembroski, H~Andrew Schwartz, and Lyle Ungar. 2017.
\newblock Recognizing counterfactual thinking in social media texts.
\newblock In \emph{Proceedings of the 55th Annual Meeting of the Association
  for Computational Linguistics (Volume 2: Short Papers)}, volume~2, pages
  654--658.

\bibitem[{Wieting et~al.(2015)Wieting, Bansal, Gimpel, and
  Livescu}]{wieting2015towards}
John Wieting, Mohit Bansal, Kevin Gimpel, and Karen Livescu. 2015.
\newblock Towards universal paraphrastic sentence embeddings.
\newblock \emph{arXiv preprint arXiv:1511.08198}.

\bibitem[{Zhang et~al.(2015)Zhang, Zhao, and LeCun}]{zhang2015character}
Xiang Zhang, Junbo Zhao, and Yann LeCun. 2015.
\newblock Character-level convolutional networks for text classification.
\newblock In \emph{Advances in neural information processing systems}, pages
  649--657.

\bibitem[{Zhou et~al.(2010)Zhou, Xu, Niu, Lan, Su, and
  Tan}]{zhou2010predicting}
Zhi-Min Zhou, Yu~Xu, Zheng-Yu Niu, Man Lan, Jian Su, and Chew~Lim Tan. 2010.
\newblock Predicting discourse connectives for implicit discourse relation
  recognition.
\newblock In \emph{Proceedings of the 23rd International Conference on
  Computational Linguistics: Posters}, pages 1507--1514. Association for
  Computational Linguistics.

\end{thebibliography}
\bibliographystyle{acl_natbib_nourl}

\end{document}